%% file: main.tex
\pdfoutput=1

\documentclass[11pt]{article}

\usepackage[preprint]{acl}

\usepackage{times}
\usepackage{latexsym}

\usepackage{graphicx}

\usepackage[T1]{fontenc}

\usepackage[utf8]{inputenc}

\usepackage{microtype}

\usepackage{inconsolata}

\usepackage{booktabs} 
\usepackage{array}
\usepackage{colortbl}
\usepackage{xcolor}
\usepackage{amssymb}
\usepackage{multirow}
\usepackage[most]{tcolorbox}
\usepackage{multicol}
\usepackage{transparent} 

\usepackage{hyperref}
\hypersetup{
    colorlinks=true,
    citecolor=[rgb]{0.2,0.3,0.4},   
    urlcolor=[rgb]{0.2,0.3,0.4},  
}

\definecolor{lightblue}{rgb}{0.93,0.95,1.0}
\definecolor{lightgray}{rgb}{0.95,0.95,0.95}

\newcolumntype{P}[1]{>{\centering\arraybackslash}p{#1}}

\title{Dynaword: From One-shot to Continuously Developed Datasets}

\author{
 \textbf{Kenneth Enevoldsen\textsuperscript{1*}},
 \textbf{Kristian Nørgaard Jensen\textsuperscript{2}},
 \textbf{Jan Kostkan\textsuperscript{1}},
 \textbf{Balázs Szabó\textsuperscript{1}},
\\
 \textbf{Márton Kardos\textsuperscript{1}},
 \textbf{Kirsten Vad\textsuperscript{1}},
\textbf{Johan Heinsen\textsuperscript{3}},
 \textbf{Andrea Blasi Núñez\textsuperscript{4}},
 \textbf{Gianluca Barmina\textsuperscript{4}},
\\
 \textbf{Jacob Nielsen\textsuperscript{4}},
 \textbf{Rasmus Larsen\textsuperscript{2}},
 \textbf{Peter Vahlstrup\textsuperscript{1}},
 \textbf{Per Møldrup Dalum\textsuperscript{1}},
 \\
 \textbf{Desmond Elliott\textsuperscript{5}},
 \textbf{Lukas Galke\textsuperscript{4}},
 \textbf{Peter Schneider-Kamp\textsuperscript{4}},
 \textbf{Kristoffer Nielbo\textsuperscript{1}}
\\
\\
 \textsuperscript{1}Aarhus University,
 \textsuperscript{2}The Alexandra Institute\\
 \textsuperscript{3}Aalborg University, \textsuperscript{4}University of Southern Denmark\\
 \textsuperscript{5}University of Copenhagen
\\ 
\\
    Correspondence: \href{mailto:Kenneth.enevoldsen@cas.au.dk}{Kenneth.enevoldsen@cas.au.dk}
}

\begin{document}
\maketitle

\begin{abstract}
Large-scale datasets are foundational for research and development in natural language processing. However, current approaches face three key challenges: (1) reliance on ambiguously licensed sources restricting use, sharing, and derivative works; (2) static dataset releases that prevent community contributions and diminish longevity; and (3) quality assurance processes restricted to publishing teams rather than leveraging community expertise.

To address these limitations, we introduce two contributions: the Dynaword approach and Danish Dynaword\footnote{Available at: \url{https://huggingface.co/datasets/danish-foundation-models/danish-dynaword}}. The Dynaword approach is a framework for creating large-scale, open datasets that can be continuously updated through community collaboration. Danish Dynaword is a concrete implementation that validates this approach and demonstrates its potential. Danish Dynaword contains over four times as many tokens as comparable releases, is exclusively openly licensed, and has received multiple contributions across industry and research. The repository includes light-weight tests to ensure data formatting, quality, and documentation, establishing a sustainable framework for ongoing community contributions and dataset evolution.
\end{abstract}
\section{Introduction}

\begin{figure}[h]
    \centering
    \includegraphics[width=1.0\linewidth]{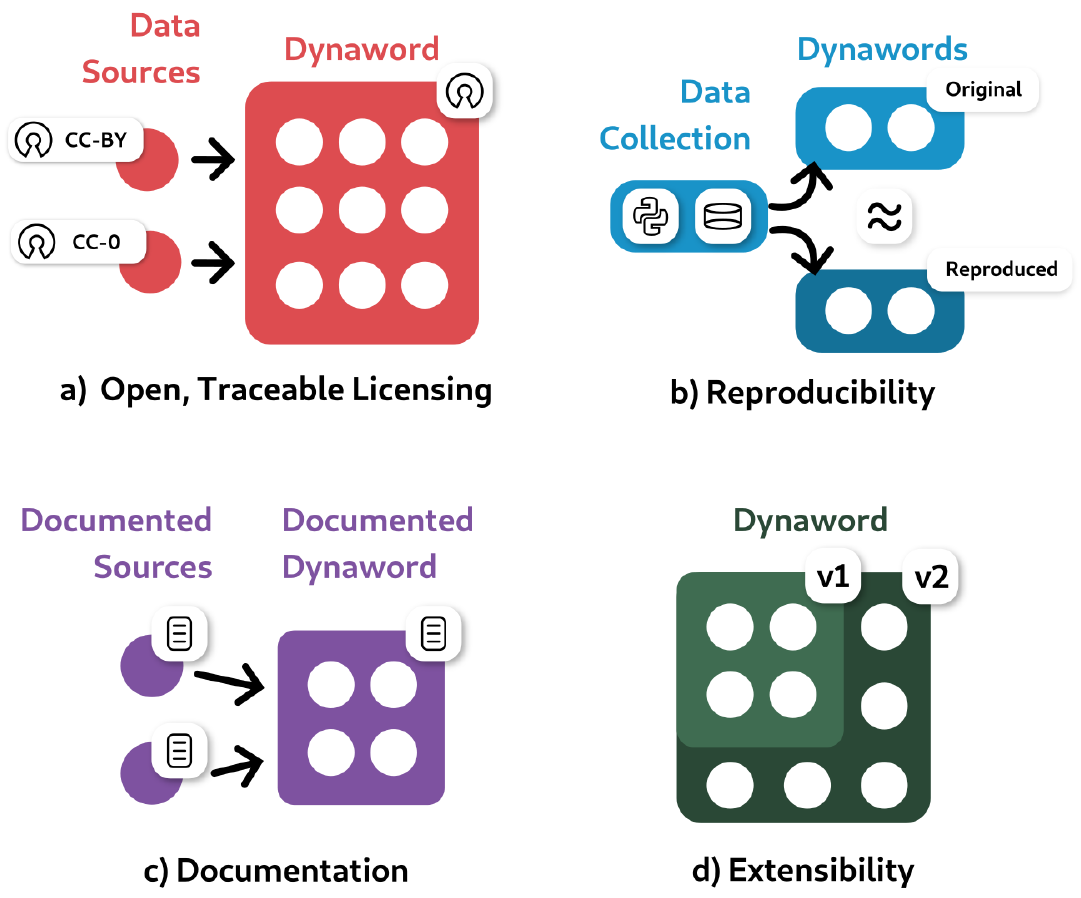}
    \caption{Overview of the guiding principles for Dynaword corpora}
    \label{fig:domains-distributions}
\end{figure}

Continuously developed open-source projects are instrumental for contemporary research and lay the foundation for the success of our fields. These projects range from foundational software such as NumPy \cite{harris2020array}, to datasets like the Universal Dependencies Treebank \cite{nivreUniversalDependenciesV12016}, to recent contributions including Flash Attention \cite{dao2022flashattention} and LoRA \cite{hu2022lora}. In open-source communities, it is understood that a project is never complete but is continually enhanced and adjusted in response to advances in the field and the software ecosystem. It is similarly recognized that relying on proprietary technology may result in legal ramifications, which, in the worst case, can render projects unusable or lead to their removal, undermining community contributions. We have already seen this impact in several instances: Udio AI Music Generator was shut down citing legal concerns; a state-of-the-art Danish encoder \cite{dfm} was removed following threats of legal action; and the Nordic Pile corpus \cite{ohman2023nordic} was never released, presumably due to copyright issues.

However, the values of open source are notably lacking for (pre-)training datasets in the field. Although we have seen large-scale releases \cite{pile, mc4}, these datasets often adhere to the pattern of being released once without updates. Even when we see continual releases \cite{fineweb, oscar}, they are frequently based on Common Crawl content. \citet{baackBestPracticesOpen2025} defines this category as \textit{open access} and notes several legal risks associated with the use of both datasets and derived models. \cite{baackBestPracticesOpen2025} also defines \textit{openly licensed} data, which enables the resharing, reuse, and modification of data, thereby providing a solid foundation for derivative works. There have been a few initiatives \cite{commoncorpus, youtube-commons}; however, despite being a step forward, these contributions still have notable shortcomings.

These datasets \cite{commoncorpus, youtube-commons} constitute a single release without reproducible code for collection. This makes it difficult to reproduce the data collection process and difficult to improve\footnote{For instance, the Common Corpus consists of multiple OCR'd documents that could likely be improved with recent advancements.} or update\footnote{E.g. the number of tokens on Wikipedia is expected to grow linearly \cite{suh2009singularity}.}. Lastly, a vague description of the underlying licenses makes it difficult to validate the license claims. For instance, stating that "Alice in Wonderland" is public domain differs from documenting that the author died in 1898, which renders it within the public domain; we refer to this as a \textit{traceable} license.

Following these limitations, we propose the \textbf{dynaword} approach for curating a corpus:

\begin{itemize}
\itemsep0em 
    \item \textbf{Traceable and open licensing}: All datasets within the corpus must be \textit{openly licensed} and maintain a \textit{traceable} license. 

    \item \textbf{Reproducibility}: It should be possible to derive a substantially similar dataset\footnote{Denoted as a 'substantially equivalent system' in the Open-Source AI definition: \url{https://opensource.org/ai/drafts/the-open-source-ai-definition-1-0-rc1}}. 

    \item \textbf{Documented}: The dataset should be well-documented under best practices in the field \cite{gebru2021datasheets}. 

    \item \textbf{Extensible}: Extending and improving the corpus should be possible, and methods for doing so should be documented. 
\end{itemize}

These guidelines are intended to provide datasets that are compatible with the open-source AI definition\footnote{\url{https://opensource.org/ai/drafts/the-open-source-ai-definition-1-0-rc1}}, FAIR \cite{wilkinson2016fair}, and compliant under a diverse set of legal frameworks, including the European Parliament Artificial Intelligence Act \cite{eu_ai_act_2024}, and conducive to creating lasting resources for both research and industry. This approach draws upon successful datasets such as the Universal Dependency Treebanks \cite{nivreUniversalDependenciesV12016}, which remain essential building blocks for linguistic analysis, model development \cite{spacy}, and recently, multilingual benchmarking of LLMs \cite{scandeval, scandeval2}.

These guidelines were developed jointly with the Danish Dynaword, which acts as a practical implementation and a testbed, showing that the guidelines are implementable and not simply ideals. We believe Danish provides an ideal case, being a low-to-mid-resource\footnote{Class 2-4 as defined by \citet{joshiStateFateLinguistic2021}} language, which has enough contributors willing to participate, yet well-contained in scope. High-resource languages\footnote{Class 5 as defined by \citet{joshiStateFateLinguistic2021}} could likely sustain multiple dynaword projects targeting specific domains such as code, academia, or healthcare.

The following sections outline how we developed a framework around Danish Dynaword that actively ensures traceable and open licensing and reproducibility, while being documented and reproducible. Our final dataset represents more than a fourfold increase in Danish tokens available compared to previous datasets and considerably improves reproducibility and documentation. We expect that this work will continue to expand with future contributions.


\section{Related Works}

\subsection{Continually released pre-training data}

Recently, dataset releases have been increasingly iterative releases with improvements to extraction, cleaning, and collection procedures. These include OSCAR \cite{oscar}, HPLT \cite{hplt, hplt2}, and fineweb \cite{fineweb}, all which are mostly based on Common Crawl\footnote{\url{https://commoncrawl.org}}. Though Common Crawl and its derivatives represent significant artifacts, the license of the underlying data remains ambiguous, raising a slew of ethical and legal concerns \cite{baackBestPracticesOpen2025}.
Additionally, these sources typically work with web-derived content, overlooking sources such as APIs, television, or radio. 

\subsection{Openly licensed data}
\label{sec:openlylicenseddata}
Data licensing is often unclear, and it is not uncommon to find datasets on Hugging Face\footnote{\url{https://huggingface.co}} published under permissive licenses containing copyrighted content. In some cases \cite{oscar, fineweb} the license applies only to the packaging, metadata, and annotations, but not to the data itself. To avoid such confusion, we will, throughout this work, utilize the three tiers defined by  \citet{baackBestPracticesOpen2025}: 1) \textbf{replicable} - enables reproduction - 2) \textit{open access} - enables download - and 3) \textit{openly licensed} - enables resharing, reuse, and modification.

Examples of openly licensed data sources intended for pre-training include YouTube Commons \cite{youtube-commons} and Common Corpus \cite{commoncorpus} as well as multiple gigaword projects \cite{graff2003english, dagw, adewumi2020corpora}. These gigaword corpora consist of ~1B (Giga) tokens available under permissive licenses. YouTube Commons compiles CC-BY licensed YouTube content, including 30B tokens of predominantly English (70\%) transcripts. The most extensive openly licensed corpus is the ~500B Common Corpus, which includes digitized newspapers and OCR'd public domain books. While these collections solved many issues, they left much to be desired. The OCR quality is often questionable, and the lack of public processing workflows makes these datasets hard to reproduce, validate and improve.


\section{Dataset Collection}

Danish Dynaword takes its outset in public segments of the Danish Gigaword \cite{dagw}, previously the largest publicly available data source for Danish. Danish Dynaword notably excludes social media data from Twitter (~32M tokens), copyrighted samples from OpenSubtitles (<1M tokens), and common crawl segments (~100M tokens) following the principles of \textit{traceable and open licensing}. We also exclude DanAvis (~30M tokens) due to a lack of overall coherence caused by scrambling.
For each source in this collection, a datasheet \cite{gebru2021datasheets} was compiled, incorporating available information, including dataset description and license references. 

Furthermore, additional open datasets were added using a process of identifying potentially openly licensed datasets, collecting them, conducting quality checks, and, lastly, reviewing their licenses in case of uncertainty. Quality checks were intentionally kept minimal to allow for downstream filtering and include verifying that the text is Danish, coherent, and readable. In the following, we will walk through each of these steps.

\textbf{Identification}: 
The majority of openly licensed datasets were found through projects such as the Danish Foundation Models \cite{dfm}, on the HuggingFace Hub, or through Sprogteknologi.dk, a website covering Danish language resources, curated by the Danish Ministry of Digitization. Additional resources were identified through social media outreach, personal communication, and issues in the \href{https://huggingface.co/datasets/danish-foundation-models/danish-dynaword}{Danish Dynaword repository}.

\textbf{License and content review}: After identification, maintainers review the data and filter out straightforward cases where the data is not Danish, is not \textit{openly licensed}, or the redistributor lacks permission to license or re-license it. For complex cases, maintainers may request legal advice from faculty services.

\textbf{Collection and Quality Checks}: 
After a dataset is checked, it is collected and undergoes quality checks. The collection procedure is documented in the form of datasheets and reproducible scripts. These scripts enable dataset updates and critical examination. Some sources excluded at this stage include the Danish subsection of Common Corpus, where the OCR was deemed insufficient (alpha ratio generally below 0.7), with most text being unreadable. If rejected, the quality issues are documented, and the issue is closed. For all sources, we deduplicate and remove short documents. 




\input{tables/comparison-table}

\subsection{Inclusion Policy}
In recent years, we have seen the rise of derived data in pre-training regimes. These data include synthetic \cite{li2023textbooks, gunasekar2023textbooks}, semi-synthetic \cite{chung2024scaling}, translations \cite{doshi-etal-2024-pretraining}, OCR'd \cite{commoncorpus}, and transcribed data \cite{youtube-commons}. These sources require deliberation before inclusion, as they can also lead to model degradation \cite{bender2021dangers, shumailov2024ai}.

While it is clear that an inclusion policy is likely to change as technologies improve, the current dynaword does not, to our knowledge, contain synthetic, machine-translated, or automatically transcribed data, but does include human-annotated audio transcriptions (e.g., FTSpeech) and translations by expert translators (e.g., Europarl) and OCR'd documents (e.g., NCC books). For OCR'd documents, we perform an additional quality check; these are described in the individual datasheets.

\subsection{Evaluation data}
It is by no mean uncommon that large public dataset include segments of evaluation data \cite{gao2020pile800gbdatasetdiverse}, this can lead overestimate of actual performance \cite{schaeffer2023pretraining, deng2024investigatingdatacontaminationmodern} and thus leading to overestimates of the model capabilities. It is therefore encourages that model developers exclude evaluation data during the training process. To facilitate this we mark datasets contained in benchmarks. For the Danish Dynaword this includes, the Danish dependency treebank, though not the annotation, which is used to create the dataset ScaLA \cite{scandeval} and Nordjyllands News, which is used for evaluating summarization \cite{scandeval, scandeval2} as well as semantic similarity \cite{seb, mmteb}.

\subsection{Contributions}

\begin{figure}
    \centering
    \includegraphics[width=1\linewidth]{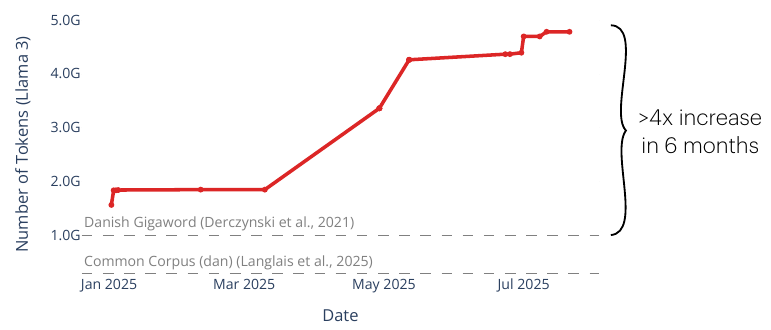}
    \caption{Number of tokens in Danish Dynaword over time.}
    \label{fig:tokens-over-time}
\end{figure}

Danish Gigaword has already seen multiple contribution prior to its official release. We show this development in \autoref{fig:tokens-over-time}. These contributions have come from companies, government institutions, individuals and universities and stemming from a wide array of background ranging for cultural heritage to NLP. 

\section{Dataset Comparison}

In \autoref{tab:size_comparison} we give a conceptual comparison of Danish Dynaword to existing openly available datasets and in \autoref{tab:trainingresults} we show the performance gap when training on Danish Dynaword instead of the Danish Gigaword, which is we eloborate on in the following section.

We present an overview of the datasets in \autoref{app:sec:dataset-overview}. For individual datasheets, we refer to the \href{https://huggingface.co/datasets/danish-foundation-models/danish-dynaword}{dataset repository}.

\subsection{Training Experiments} 
To study relative quality Danish Dynaword and Danish Gigaword~\cite{stromberg-derczynski-etal-2021-danish} we perform set of training experiments to estimate the expected language modelling performance.

The experiments were performed using the Gemma-1B model either continually pre-trained\footnote{Using the \href{https://huggingface.co/google/gemma-3-1b-pt/tree/fcf18a2a879aab110ca39f8bffbccd5d49d8eb29}{gemma-3-1b-pt} checkpoint.} or trained from scratch. To ensure a fair comparison we train two models on Dynaword, one \textit{matched} in size to Danish Gigaword and one trained on the \textit{full} dataset. 

We evaluated the perplexity performance on four datasets from Dynaword: DDT, JVJ, Synnejysk.dk, and Nordjyllands News which were held out during training. Additionally, we tested on contemporary sources not included in Dynaword: news articles from DR (dr.dk) and Danish Wikipedia articles published after January 1, 2025.

All models were trained with a maximum sequence length of 6144 tokens, an effective batch size of 32 (via gradient accumulation), and an initial learning rate of $10^{-5}$ for pretrained models and $10^{-3}$ for the models from scratch, in both cases we use a cosine learning rate scheduler. Models were trained using the Danish Dynaword version 1.2.0 and with training code available on GitHub\footnote{\url{https://github.com/schneiderkamplab/offpolicy_kd/}, Commit 76b546e}. An overview of models is provided in Appendix~\ref{app:models}.

\input{tables/lm-comparison}

Table~\ref{tab:trainingresults} shows the perplexity across the six held-out datasets. 
Comparing Dynaword to Gigaword, we see an average  relative improvement of 5.9\% for continual pre-training starting from Gemma-3-1b-pt, and an improvement of 26\% for Gemma-3-1b models trained from scratch.
Even in the size-matched scenario, Dynaword yields improvements of 2.6\% with continual pre-training and 18\% when training from scratch.
Results from a downstream evaluation on Danish tasks from EuroEval~\cite{scandeval} show that continual pre-training on Danish Dynaword yields improvements on 7 out of 9 tasks. Detailed results are provided in Appendix~\ref{app:downstream}.

\section{Conclusions}

In this work, we argue for continuously developed datasets. We dub such datasets dynawords and outline four key principles which are required for continuous development; a) Open and traceable licensing, b) Reproducibility, c) Documentation, and d) Extensibility. As a testbed and concrete implementation, we release Danish Dynaword, the largest \textit{openly licensed} Danish corpus, which we expect to grow with future submissions. 
Danish Dynaword has already received contributions from multiple parties, including industry, private individuals, and research. For the datasets in the dynaword, we also provide reproducible scripts to update them and a compliance review of the licensing. 

To enable future contributions, Danish Dynaword comes with lightweight tests to ensure formatting, documentation, and dataset quality. Danish Dynaword has already seen contributions from multiple parties, including industry, private individuals, and research.

While Dynaword remains an order of magnitude smaller than non-openly licensed sources, it provides a sustainable foundation for building models. Dynawords should be seen as a complement to existing efforts, making high-quality and permissible data available. We hope that this dynaword can be a blueprint for future dynawords targeting other languages or domains.
 
\section*{Limitations}
\textbf{Size} 
Danish Dynaword significantly expands available \textit{openly-licensed} data, with future growth expected from initiatives like Dansk Sprogmodel Konsortium (DSK)\footnote{\url{https://alexandra.dk/dsk/}} and national AI programs\footnote{\url{https://digst.dk/strategier/strategi-for-kunstig-intelligens/}}. However, it remains an order of magnitude smaller than the Common Crawl datasets \citep{fineweb, oscar}. This gap may persist, but could be addressed through multilingual or multimodal sources.

\noindent\textbf{Coverage and bias}:
Danish Dynaword, given its requirements, is biased toward domains with clear licensing. Thus, the data set only contains limited amounts of social media data and a disproportionate amount of legal documents. You can explore the coverage by domain in \autoref{fig:domains-distributions}.

\begin{figure}[h!]
    \centering
    \includegraphics[width=0.8\linewidth]{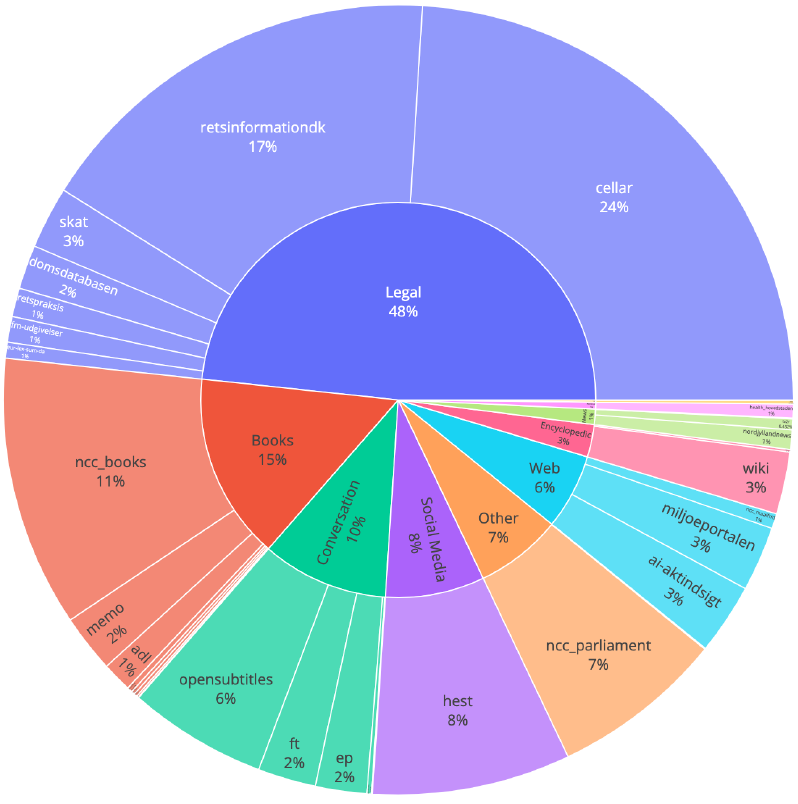}
    \caption{Content by domain. The inner circle shows the domain, while the outer layers are the source.}
    \label{fig:domains-distributions}
\end{figure}

\noindent\textbf{Only Danish}:
Dynaword presents a methodology for developing large-scale public language resources. However, in this work, we only present one such resource, namely, for Danish. We hope that Danish Dynaword can act as a starting point for similar efforts for other languages.

\textbf{Review quality and dataset poisoning}:
Throughout the development process, it became clear that contributing minor changes, such as filtering out a few bad examples, was difficult, both due to the limited support for reviewing large data changes. 
While previous projects (e.g. \citet{nivreUniversalDependenciesV12016}) have tackled this issue using human-readable formats, this is likely inefficient at the current scale.

This lack of clarity increased the likelihood of dataset attacks such as dataset poisoning \cite{goldblum2022dataset}. We expect to see both interface development and software development to detect and prevent such attacks and ensure review quality.



\subsection*{Ethical consideration}
Despite our effort to prevent issues, large-scale dataset development often involves seemingly \textit{openly-licensed} datasets that may contain copyrighted content. A notable instance occurred with the initial release of OpenSubtitles, which was part of the Danish Gigaword \cite{dagw}. By providing a versioned dataset along with a changelog, we clearly indicate when sources are excluded, thereby promoting transparency.


\section*{Acknowledgments}
 
Part of the computation done for this project was performed on the UCloud interactive HPC system, which is managed by the eScience Center at the University of Southern Denmark. The Danish e-Infrastructure Consortium (DeIC) granted the compute resources (DeiC-AU-N1-2025144 and DeiC-AU-N1-2025118).
This project was made possible by funding from the Danish Government to Danish Foundation Models (4378-00001B).
A special thanks goes to organizations that have publicly made their data available under openly licensed regimes and to individuals and organizations that have encouraged this development.

\bibliography{references}

\newpage
\input{appendix}

\end{document}

%% file: tables/comparison-table.tex
\begin{table*}
\resizebox{\textwidth}{!}{
\centering
\small
\begin{tabular}{llp{1.8cm}p{1.4cm}p{1.8cm}p{2.2cm}p{0cm}}
\hline
 &  &  &  & \multicolumn{3}{c}{\textbf{Tiers of Dataset Openness}} \\
\cline{5-7}
\textbf{Dataset} & \textbf{Size} & \textbf{Contributions}  & \textbf{Replicable} & \textbf{Open Access} & \textbf{Openly Licensed}\\
 & Tokens & Is there a clear process for contributions & Enables the user to reproduce. & Enables the user to download. & Enables the user to reuse, share, and modify. \\
\hline
\multicolumn{6}{l}{\textbf{Tier 3: Openly Licensed}} \\
\rowcolor[gray]{0.9} \hspace{0.5em}Danish Dynaword (v1.2.7) & \phantom{$\sim$}4.8B  & \centering Yes & \centering\checkmark& \centering\checkmark& \centering\checkmark& \\
\hspace{0.5em}Danish Gigaword \cite{dagw} & $\sim$1B  &  \centering No & & \centering\checkmark& \centering\checkmark& \\
    \hspace{0.5em}Common Corpus (dan) \cite{commoncorpus} & $\sim$0.3B  & \centering No & & \centering\checkmark & \centering\checkmark& \\
\hline
\multicolumn{6}{l}{\textbf{Tier 2: Open Access}} \\
\hspace{0.5em}SnakModel \cite{snakmodel}  & $\sim$13.6B  & \centering No & \centering\checkmark& \centering\checkmark& \\
\hspace{0.5em}Fineweb (dan) \cite{fineweb} & $\sim$26B & \centering (Yes)$^1$ & \centering\checkmark& \centering\checkmark& \\
\hline
\end{tabular}
}
\caption{Comparison of Danish Language Datasets. Danish Gigaword only includes segments that are currently publicly available. \\ $^1$Has a Discord channel and encourages involvement.}
\label{tab:size_comparison}

\end{table*}

%% file: tables/lm-comparison.tex
\begin{table*}[]
    \small
    \centering
    \begin{tabular}{lllllll}
    \toprule
        & \multicolumn{4}{c}{\transparent{0.7}Held out Datasets} & \multicolumn{2}{c}{\transparent{0.7}Contemporary (2025)} \\
        \cmidrule(lr){2-5}\cmidrule(lr){6-7} 
         & \textbf{Treebank} & \textbf{Fiction} & \textbf{Dialect} & \textbf{News} & \textbf{Wikipedia} & \textbf{News} \\
         & DDT & JVJ & Synnejysk & Nordjylland & Wiki (dan) & DR  \\
         \midrule
    \textbf{Reference baseline} \\
         \hspace{3mm}Gemma-3-1b-pt & 16.0 & 33.9 & 62.8 & 9.8 & 9.2 & 9.7 \\
    \textbf{Continual pre-training} \\
         \hspace{3mm}Gigaword* & 14.2 & 29.1 & 54.1 & 8.5 & 8.1 & 9.2 \\
         \hspace{3mm}Dynaword* (matched) & 14.0 (+1.2\%) & 26.6 (+8.5\%) & 51.8 (+4.3\%) & 8.4 (+1.2\%) & 8.2 (-0.7\%) & 9.1 (+0.9\%) \\
         \hspace{3mm}Dynaword* (full) & 13.5 (+4.6\%) & 25.2 (+13\%) & 50.1 (+7.5\%) & 8.1 (+4.6\%) & 8.0  (+1.9\%) & 8.9 (+3.5\%) \\ 
    \textbf{Pre-training from scratch}\\
         \hspace{3mm}Gigaword* & 48.8           & 128           & 219 & 24.4 & 26.7  & 29.4 \\
         \hspace{3mm}Dynaword* (matched) & 43.0 (+12\%) & 86.2 (+33\%) & 144 (+34\%) & 21.0 (+14\%) & 26.0 (+3.7\%) & 25.4 (+14\%)\\
         \hspace{3mm}Dynaword* (full)         & 39.2 (+20\%) & 79.4 (+38\%) & 128 (+42\%) & 19.6 (+21\%) & 23.2 (+13\%) & 23.6 (+20\%) \\
         \bottomrule
    \end{tabular}
    \caption{Perplexity for Gemma-3-1b models continually pre-trained (middle) and pre-trained from scratch (bottom) on Gigaword, Dynaword (size-matched to Gigaword), and the full Dynaword dataset. Relative performance is calculated with respect to the Gigaword baseline. *The four validation datasets were excluded from the training data.}
    \label{tab:trainingresults}
\end{table*}

%% file: appendix.tex
\appendix
\onecolumn

\section{Author Contributions}
\label{sec:appendix}

Table \ref{app:tab:author-contributions} shows authorship contributions across different categories. All authors have agreed to the final version of the print.

\input{tables/Appendix-author-contributions}

\section{Dataset Overview}
\label{app:sec:dataset-overview}

\autoref{tab:dataset_overview} shows the overview of datasets within Danish Dynaword, including a short description, license, and size. For an updated view, we recommend checking out the Huggingface repository\footnote{\url{https://huggingface.co/datasets/danish-foundation-models/danish-dynaword}}

\input{tables/dataset-overview}

\begin{table}[h]
\centering
\begin{tabular}{lll}
\toprule
\textbf{Model name} & \textbf{checkpoint} & \textbf{trained on} \\
\midrule
gemma-3-1b-cpt-gigaword-v1 & gemma-3-1b-pt & Danish Gigaword* \\

 gemma-3-1b-cpt-dynaword-matched-v1 & gemma-3-1b-pt & Danish Dynaword* (matched) \\

 gemma-3-1b-cpt-dynaword-full-v1
 & gemma-3-1b-pt & Danish Dynaword* (full) \\
gemma-3-1b-scratch-gigaword-v1 & random init. & Danish Gigaword*  \\

gemma-3-1b-scratch-dynaword-matched-v1 & random init. & Danish Dynaword* (matched) \\

gemma-3-1b-scratch-dynaword-full-v1 & random init. & Danish Dynaword* (full) \\

\bottomrule
\end{tabular}
\caption{Overview of trained models. 
*The four validation datasets (DDT, JVJ, Synnejysk, Nordjylland) were excluded from the training data.}\label{app:tab:models}
\end{table}

\section{Trained Models}\label{app:models}
Table~\ref{app:tab:models} provides an overview of the trained models for the dataset comparison including access links. Despite models are limited in size to 1B, the results show that Danish Dynaword yields gains on 7 out of 9 downstream tasks.

\section{Downstream Evaluation}\label{app:downstream}

Tables~\ref{tab:euroeval:1} and \ref{tab:euroeval:2} report the scores of the continually pre-trained models on the Danish part of the EuroEval benchmark~\cite{scandeval}. Results show the expected gains

\input{tables/euroeval_nlu}

\input{tables/euroeval_nlg}

%% file: tables/Appendix-author-contributions.tex
\begin{table}[h]
\centering
\begin{tabular}{P{5cm}P{5cm}P{5cm}}
\toprule
\textbf{Conceptualization}  & \textbf{Writing} & \textbf{Writing} \\
\textit{Idea, narrative, planning} & \textit{Original draft} & \textit{Review and editing} \\
\addlinespace[2pt]
Kenneth Enevoldsen & Kenneth Enevoldsen & Kenneth Enevoldsen \\
Jan Kostkan &  &  Márton Kardos \\
Desmond Elliott &  & Kristoffer Nielbo \\
Kristoffer Nielbo &  & Per Møldrum Dalum \\
Kristian Nørgaard Jensen & & \\
\addlinespace[10pt]
\textbf{Data curation} & \textbf{Data review} & \textbf{Software} \\
\textit{Download, select, process} & \textit{Licensing} &  \textit{CI, testing} \\
\addlinespace[2pt]
Kenneth Enevoldsen &  Per Møldrum Dalum & Kenneth Enevoldsen \\
Kristian Nørgaard Jensen &  & Kristian Nørgaard Jensen \\
Jan Kostkan &  &   \\
Balázs Szabó & &  \\
Peter Vahlstrup & & \\
\addlinespace[10pt]
\textbf{Dataset Comparison} & \textbf{Dataset Contributors} & \textbf{Supervision and Funding} \\
\textit{Model training, evaluation} & \textit{Curation, annotation, documentation} & \textit{Ideation, planning}   \\
\addlinespace[2pt]
Andrea Blasi Núñez & Kirsten Vad & Peter Schneider-Kamp \\
Gianluca Barmina &  Johan Heinsen & Kristoffer Nielbo \\
Jacob Nielsen &  &  \\
Rasmus Larsen &  &  \\
Lukas Galke &  &  \\
Peter Schneider-Kamp &  &  \\
\addlinespace[5pt]
\bottomrule
\end{tabular}
\caption{Paper contributions.}
\label{app:tab:author-contributions}
\end{table}

%% file: tables/dataset-overview.tex
\begin{table*}
    \centering
\begin{tabular}{llll}
\toprule
\textbf{Source} & \textbf{Description} & \textbf{Size} & \textbf{License} \\
\midrule
\textbf{Legal} \\
\hspace{0.5em}Cellar & EU legal documents and open data & 1.15B & CC-BY-SA 4.0 \\
\hspace{0.5em}retsinformation.dk & The legal information system of Denmark & 818.25M & Copyright Law \\
\hspace{0.5em}Skat.dk & The Danish tax authority website & 122.11M & CC-0 \\
\hspace{0.5em}Domsdatabasen.dk & Selected judgments from the Danish courts & 86.35M & Copyright Law \\
\hspace{0.5em}Retspraksis & Case law or judicial practice in Denmark & 56.26M & CC-0 \\
\hspace{0.5em}EUR-lex SUM & EU legislation with summaries & 31.37M & CC-BY-SA 4.0 \\
\textbf{Social Media} \\
\hspace{0.5em}Heste-nettet.dk & Danish Debate forum  & 389.32M & CC-0 \\
\textbf{Spoken} \\
\hspace{0.5em}Opensubtitles & Movie Subtitles from OpenSubtitles & 271.60M & CC-0 \\
\hspace{0.5em}FT & Meeting records from the Danish parliament & 114.09M & CC-0 \\
\hspace{0.5em}Danske Taler & Speeches from dansketaler.dk & 8.72M & CC-0 \\
\hspace{0.5em}Spont & Conversational samples from research project & 1.56M & CC-0 \\
\hspace{0.5em}NAAT & Danish speeches from 1930-2022 & 286.68K & CC-0 \\
\textbf{Web} \\
\hspace{0.5em} Municipal Websites & Municipality websites from AI-aktindsigt & 139.23M & Apache 2.0 \\
\hspace{0.5em}Miljoeportalen & Environmental Reports from Miljøportalen & 127.38M & CC-0 \\
\hspace{0.5em}FM Udgivelser & Publication of the Ministry of Finance & 50.34M & CC-BY-SA 4.0 \\
\hspace{0.5em}NCC Maalfrid & Danish content from Norwegian institutions & 29.26M & NLOD 2.0 \\
\textbf{Medical} \\
\hspace{0.5em}Health Hovedstaden & Guidelines and info. documents for healthcare & 27.07M & CC-0 \\
\textbf{Encyclopedic} \\
\hspace{0.5em}Wikipedia & The Danish subsection of Wikipedia & 122.00M & CC-0 \\
\hspace{0.5em}Europarl & The Danish subsection of Europarl & 100.84M & CC-0 \\
\hspace{0.5em}WikiSource & The Danish subsection of Wikisource & 5.34M & CC-0 \\
\textbf{Books and Novels} \\
\hspace{0.5em}NCC Books &  OCR'ed Danish books & 531.97M & CC-0 \\
\hspace{0.5em}MeMo & Novels from the Modern Breakthrough & 113.74M & CC-BY-SA 4.0 \\
\hspace{0.5em}ADL & Danish literature from 1700-2023 & 58.49M & CC-0 \\
\hspace{0.5em}Grundtvig & The complete works of N.F.S. Grundtvig & 10.53M & CC-0 \\
\hspace{0.5em}Gutenberg & Books from Project Gutenberg & 6.76M & Gutenberg \\
\hspace{0.5em}WikiBooks & The Danish Subsection of Wikibooks & 6.24M & CC-0 \\
\hspace{0.5em}JVJ & The works of Johannes V. Jensen & 3.55M & CC-BY-SA 4.0 \\
\textbf{News}\\
\hspace{0.5em}Nordjylland News & Articles from Newspaper TV2 Nord & 37.90M & CC-0 \\
\hspace{0.5em}TV2R & Articles from TV2R & 21.67M & CC-BY-SA 4.0 \\
\hspace{0.5em}NCC Newspapers & OCR'd Newwspapers from NCC & 1.057M & CC-0 \\
\textbf{Dialect}\\
\hspace{0.5em}Botxt & Dictionary of the dialect Bornholmsk & 847.97K & CC-0 \\
\hspace{0.5em}Synnejysk.dk & Dataset of the dialect Sønderjysk & 52.02K & CC-0 \\
\textbf{Other}\\
\hspace{0.5em}NCC Parliament & OCR'ed Danish from the Norwegian parliament & 338.87M & NLOD 2.0 \\
\hspace{0.5em}Nota & The text segment from readaloud data & 7.30M & CC-0 \\
\hspace{0.5em}DanNet & A Danish WordNet & 1.48M & DanNet 1.0 \\
\hspace{0.5em}Religious texts & Religious text from the 1700-2022 & 1.24M & CC-0 \\
\hspace{0.5em}DDT & The Danish Dependency Treebank & 185.45K & CC-BY-SA 4.0 \\
\midrule
\textbf{Total} &  & 4.80B &  \\
\bottomrule
\end{tabular}
    \caption{Overview of the dataset in Danish Dynaword (v1.2.7). Size in Llama 3 tokens.}
    \label{tab:dataset_overview}
\end{table*}

%% file: tables/euroeval_nlu.tex
\begin{table}[]
    \footnotesize
    \centering
    \makebox[\textwidth]{
        \begin{tabular}{lcccccc}
        \toprule
            & \textbf{angry-tweets} & \textbf{dansk} & \textbf{scandiqa-da} & \textbf{da-talemaader} & \textbf{da-citizen-tests}  \\
            \textcolor{gray}{Task ($\rightarrow$)} & \textcolor{gray}{Sentiment}  & \textcolor{gray}{NER} & \textcolor{gray}{Reading comp.} & \textcolor{gray}{Knowledge} & \textcolor{gray}{Knowledge}   \\
            \textcolor{gray}{Score ($\rightarrow$)} & \textcolor{gray}{(MCC)} & \textcolor{gray}{(Micro F1)} & \textcolor{gray}{(F1)} & \textcolor{gray}{(Accuracy)} & \textcolor{gray}{(Accuracy)} \\
            \midrule
            \textbf{Reference baseline}\\
            \hspace{3mm}Gemma-3-1b-pt & 37.27 $\pm$ 2.28 & 14.55 $\pm$ 0.86 & 51.80 $\pm$ 3.95 & 23.28 $\pm$ 4.58 & 40.78 $\pm$ 3.28 \\
            \textbf{Continual Pre-training}\\
            \hspace{3mm}Gigaword* & 36.58 $\pm$ 3.04 & 15.19 $\pm$ 1.47 & 48.10 $\pm$ 1.46 & 25.47 $\pm$ 6.19 & 39.67 $\pm$ 4.03                       \\
            \hspace{3mm}Dynaword* (matched) & \textbf{38.80} $\pm$ 1.84 & \textbf{17.97} $\pm$ 1.79 & \textbf{50.04} $\pm$ 3.33 & \textbf{35.31} $\pm$ 5.41 & 35.22 $\pm$ 4.00      \\
            \hspace{3mm}Dynaword* (full) & \textbf{38.80} $\pm$ 1.92 & \textbf{16.30} $\pm$ 2.43 & \textbf{49.07} $\pm$ 3.43 & \textbf{32.19} $\pm$ 4.04 & 36.22 $\pm$ 3.23    \\
            \bottomrule
        \end{tabular}
    }
    \caption{Downstream results on the Danish subsection of EuroEval(NLU) of Gemma-3-1b models continually pre-trained (bottom) on Gigaword, Dynaword (size-matched to Gigaword), and the full Dynaword dataset. Scores are reported alongside standard error of the mean.
    Values marked in bold indicate an increase over the Gigaword baseline. *All the evaluation datasets were excluded from all training runs.}
    \label{tab:euroeval:1}
\end{table}

%% file: tables/euroeval_nlg.tex
\begin{table}[]
    \footnotesize
    \centering
    \makebox[\textwidth]{
        \begin{tabular}{lllcccc}
        \toprule
             & \multicolumn{2}{c}{\textbf{nordjylland-news}} & \textbf{hellaswag-da} & \textbf{scala-da}  \\
            \textcolor{gray}{Task ($\rightarrow$)} & \multicolumn{2}{c}{\textcolor{gray}{Summarization}} & \textcolor{gray}{Common sense} & \textcolor{gray}{Linguistic acceptability} \\
            \textcolor{gray}{Score ($\rightarrow$)} & \textcolor{gray}{(BertScore)} & \textcolor{gray}{(Rouge-L)} & \textcolor{gray}{(Accuracy)} & \textcolor{gray}{(MCC)} \\
            \midrule
            \textbf{Reference baseline}\\
            \hspace{3mm}Gemma-3-1b-pt & 56.93 $\pm$ 1.76 & 11.31 $\pm$ 1.60  & 24.61 $\pm$ 2.19 & 0.84 $\pm$ 4.43\\
            \midrule
            \textbf{Continual Pre-training}\\
            \hspace{3mm}Gigaword* & 47.36 $\pm$ 3.97 & 7.77 $\pm$ 0.72 & 24.77 $\pm$ 2.10 & 1.27 $\pm$ 6.79                     \\
            \hspace{3mm}Dynaword* (matched) & 46.40 $\pm$ 3.00 & 7.59 $\pm$ 0.74 & \textbf{26.84} $\pm$ 2.23 & 0.74 $\pm$ 3.85    \\
            \hspace{3mm}Dynaword* (full) & 47.06 $\pm$ 2.86 & \textbf{7.84} $\pm$ 0.69 & \textbf{24.92}	$\pm$ 2.28 & \textbf{1.86 }$\pm$ 3.66     \\
            \bottomrule
        \end{tabular}
    }
    \caption{Downstream results on the Danish subsection of EuroEval(NLG) of Gemma-3-1b models continually pre-trained (bottom) on Gigaword, Dynaword (size-matched to Gigaword), and the full Dynaword dataset. Scores are reported alongside standard error of the mean. Values marked in bold indicate an increase over the Gigaword baseline. *All the evaluation datasets were excluded from all training runs.}
    \label{tab:euroeval:2}
\end{table}